\documentclass[letterpaper, 10 pt, conference]{ieeeconf}  

\IEEEoverridecommandlockouts                              

\usepackage{graphicx}
\usepackage{listings}
\usepackage{xcolor}

\lstdefinelanguage{json}{
    basicstyle=\ttfamily\scriptsize,
    breaklines=true,
    frame=lines,
    backgroundcolor=\color{gray!10},
    showstringspaces=false,
    string=[db]{"},
    stringstyle=\color{green!50!black},
    morestring=[s][\color{blue}]{\ \ "}{":},
    keywordstyle=\color{blue},
    keywords={true,false,null},
    literate=
     *{0}{{{\color{red}0}}}{1}
      {1}{{{\color{red}1}}}{1}
      {2}{{{\color{red}2}}}{1}
      {3}{{{\color{red}3}}}{1}
      {4}{{{\color{red}4}}}{1}
      {5}{{{\color{red}5}}}{1}
      {6}{{{\color{red}6}}}{1}
      {7}{{{\color{red}7}}}{1}
      {8}{{{\color{red}8}}}{1}
      {9}{{{\color{red}9}}}{1}
      {.}{{{\color{red}.}}}{1}
      {:}{{{\color{gray}{:}}}}{1}
      {,}{{{\color{gray}{,}}}}{1}
      {\{}{{{\color{gray}{\{}}}}{1}
      {\}}{{{\color{gray}{\}}}}}{1}
      {[}{{{\color{gray}{[}}}}{1}
      {]}{{{\color{gray}{]}}}}{1},
}

\title{\LARGE \bf
Prompt2Craft: Generating Functional Craft Assemblies with LLMsarg
}

\author{Vitor Hideyo Isume$^{1}$, Takuya Kiyokawa$^{1}$, Natsuki Yamanobe$^{2}$, Yukiyasu Domae$^{2}$, Weiwei Wan$^{1}$\\ and Kensuke Harada$^{1, 2}$%
\thanks{$^{1}$Vitor Hideyo Isume, Takuya Kiyokawa, Weiwei Wan and Kensuke Harada are with the Graduate School of Engineering Science,
Osaka University, Osaka, Japan {\tt\small isume@hlab.sys.es.osaka-u.ac.jp, \{kiyokawa, wan, harada\}@sys.es.osaka-u.ac.jp}}%
\thanks{$^{2}$Natsuki Yamanobe, Yukiyasu Domae and Kensuke Harada are with the National Institute of Advanced Industrial Science and Technology (AIST), Tokyo, Japan {\tt\small \{n-yamanobe, domae.yukiyasu\}@aist.go.jp}}%
}

\begin{document}

\maketitle
\thispagestyle{empty}
\pagestyle{empty}

\begin{abstract}

The Craft Assembly Task - a robotic assembly task inspired by handmade crafts - poses unique challenges relative to a traditional object assembly task. It involves open-ended design decisions that are difficult to automate, with previous work relying on prior assumptions or expert knowledge. In this work, we propose to employ an off-the-shelf Large Language Model~(LLM) as the main decision maker to autonomously choose a selection of objects, their positions and connections without additional training or fine-tuning, using an RGB image of the target object in the wild as a reference. At the core of our method is a customized structure for the assembly task description to constrain the LLM's predictions. The proposals are then verified through collision checks and physics simulation. To enable moving components for the craft, we allow boolean operations between parts to create slots for them, if necessary. We evaluate our approach on eight types of objects with three possible functions: "hit", "support" and "rolling". For visual similarity, we compare our generated final crafts with the 3D models generated by a state-of-the-art image-to-3D generation foundation model, and a baseline that uses an image-to-3D part generation model to directly determine the parts. Our approach consistently produces coherent, physically plausible assemblies, while the baseline often fails to generate or segment the object into parts suitable for assembly.

\end{abstract}

\section{Introduction}

The Craft Assembly Task~\cite{isume2024component} is defined as a type of assembly task inspired by handmade crafts. Given the RGB image of a target object in the wild and a list of the available objects in the scene, the goal is to generate a craft that is similar to the input image. Compared to a typical object assembly, this task is under-specified, as the available objects are not exact correspondences to parts of a possible solution, and the final craft will not be an exact representation of the given RGB image, as they may differ due to the appearance or limitations of the available objects for the assembly. In previous work, specific object class data was prepared beforehand for template matching and to fine-tune networks, which limited their scalability and robustness. 

Meanwhile, Large Language Models~(LLMs) have demonstrated the ability to interpret natural language commands and ground them to possible actions. As the craft assembly involves abstract reasoning, it naturally aligns with the strengths of LLMs, making them well-suited as decision making agents for this task. However, although they achieved impressive results for high-level planning, most LLMs struggle in spatial reasoning~\cite{cheng2024spatialrgpt}, and complex reasoning tasks~\cite{guo2025r}, with recent models implementing specific techniques to improve reasoning~\cite{jaech2024openai}.

\begin{figure}
    \centering    \includegraphics[width=\columnwidth]{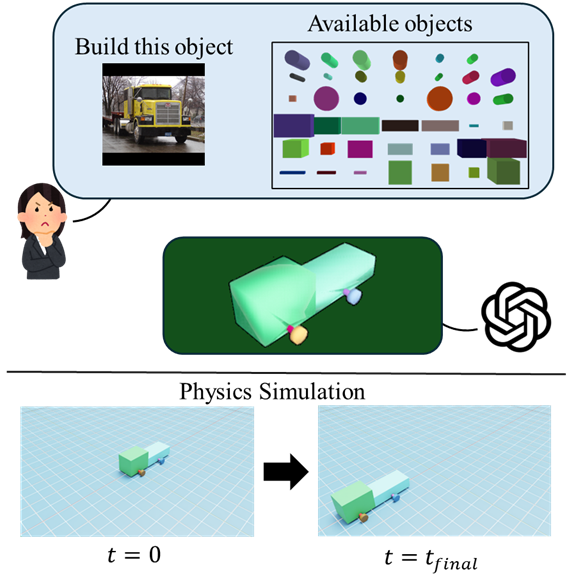}
    \caption{Concept of our method on using LLMs to generate craft assemblies which are validated in physics simulation for functionality.}
    \label{fig:example_intro}
\end{figure}

In this work, we leverage off-the-shelf multi-modal LLMs as the main decision-making agent for the craft assembly task, without additional training or fine-tuning. The LLM agent is tasked with inferring the list of parts, selecting suitable objects from the available set, determining their poses, how they are connected and if modifications are required. To address the limitations of current models in spatial reasoning, we design a structured representation of the craft assembly that limits the parameters the LLM must predict. Beyond the visual characteristics of the craft, it must also be able to perform a target function. In the previous work, additional parts for functionality were added through a rule-based approach. In contrast, to support moving components, we allow the agent to perform boolean operations to generate slots between parts when necessary. 

The generated responses by the LLM are validated through format checking, collision detection and physics-based functional testing. If any error is detected, the agent can be re-prompted up to two times. We evaluate our approach on eight object categories - hammer, bookshelf, chair, table, bus, scooter, skateboard, truck - across three possible functions  - hit, support and rolling -, under one scene with 41 types of available objects for the assemblies, composed of either cuboids and cylinders with varying dimensions. To evaluate visual similarity, due to a lack of ground truth 3D models for the inputs, we use a state-of-the-art image-to-3D model, TripoSG~\cite{li2025triposg}, to generate meshes that serve as an approximated ground truth. Then we compare the results of our approach to a baseline, generated by using a state-of-the-art image-to-3D parts model, PartCrafter~\cite{lin2025partcrafter}, and combining it with the primitive simplification and object selection from~\cite{isume2024component}. Our method achieves a high success rate in generating stable and functional assemblies, while maintaining visually coherent results compared to the baseline.

\section{Related Works}

\subsection{Part Assembly}

Autonomous part assembly is a complex task involving pose estimation of the parts, sequence planning and contact-rich interactions. Many works that focus on the pose estimation step explored the usage of different representations of the target to guide the assembly, such as point clouds~\cite{li2023rearrangement}, images~\cite{li2020learning} or intermediate representations such as instruction manuals~\cite{zhang2024manual}~\cite{tie2025manual2skill}. Alternatively, other works explored combinatorial exploration~\cite{Li_2024_CVPR}~\cite{xu2024spaformer} to generate complete assemblies directly from the given parts, without explicit target supervision.


These works usually assume that all provided parts will be used, and focus on the final representation accuracy, without considering the connection and collision between the parts, as the parts representation often do not contain details about the joints or connection points, except in cases where their CAD files are available~\cite{Willis_2022_CVPR}. In the Craft Assembly Task, part segmentation and selection are unknown and must be inferred. In this work, we also consider collision and connection types between parts to properly validate the craft's stability and functionality in a physics simulation.

\subsection{Large Language Models in Task Planning}

Other works exploring the usage of LLM agents in robotics mainly focus on general task planning, where those agents play a key role to ground natural language instructions into known primitive actions, facilitating non-expert users to interact and use specialized autonomous systems~\cite{brohan2023can}. However, a common weakness of LLMs is their limited capability for solving complex tasks or numerical calculations~\cite{rae2021scaling}. Specific prompting strategies to elicit reasoning, such as chain-of-thought~\cite{wei2022chain} and error handling strategies~\cite{wang2024llmˆ}, have been proposed to handle more complex robotic tasks. Another effective strategy is few-shot prompting, where successful and failure examples are provided to generate new robotic tasks ~\cite{wang2023gensim}~\cite{mower2024ros}. Recent models were further trained using large-scale reinforcement learning on chain-of-thought~\cite{jaech2024openai} to perform better in complex reasoning tasks.

In assembly tasks, LLM agents are mainly used for high-level planning by generating a sequence of sub-tasks, typically API calls, to accomplish the task~\cite{macaluso2024toward}, with precise numerical information being handled by outside sensors and actions. In this work, instead of a sequence of actions, we explore the usage of LLM agents for direct inference of 3D geometric relationships among parts, such as orientation, connectivity and relative position.

\subsection{Creative Assembly Tasks}

In the context of robotic assembly, some works have explored the following setting: given a non-exact representation of a target object, such as a 2D sketch~\cite{xu2025stack}, or a semantic target word~\cite{goldberg2025blox}, and a set of available objects, they need to select a subset of those objects and assemble the target. However, they are limited to stacking operations. In contrast, our work considers the usage of LLM in more complex scenarios, by allowing parts to be positioned relative to each other in six orthogonal positions (front, back, right, left, above, below) and connected via two types of joints: fixed, where parts are attached together, and non-fixed, where they are free-standing, interlocked through geometric constraints. 

The original proposal for the Craft Assembly Task~\cite{isume2024component} had limitations in the variability and scalability of the generated results, as the list of parts and their dimensions depended on predefined templates and rule-based mechanisms for functionality. We address these limitations by allowing a multimodal LLM agent to directly infer the object and a list of parts from the input RGB image of the object in the wild, without any additional training, and propose the mechanism for a target function, validating it through physics simulation.

\section{Methodology}

\begin{figure*}[t]
    \smallskip
    \centering
    \includegraphics[width=\linewidth]{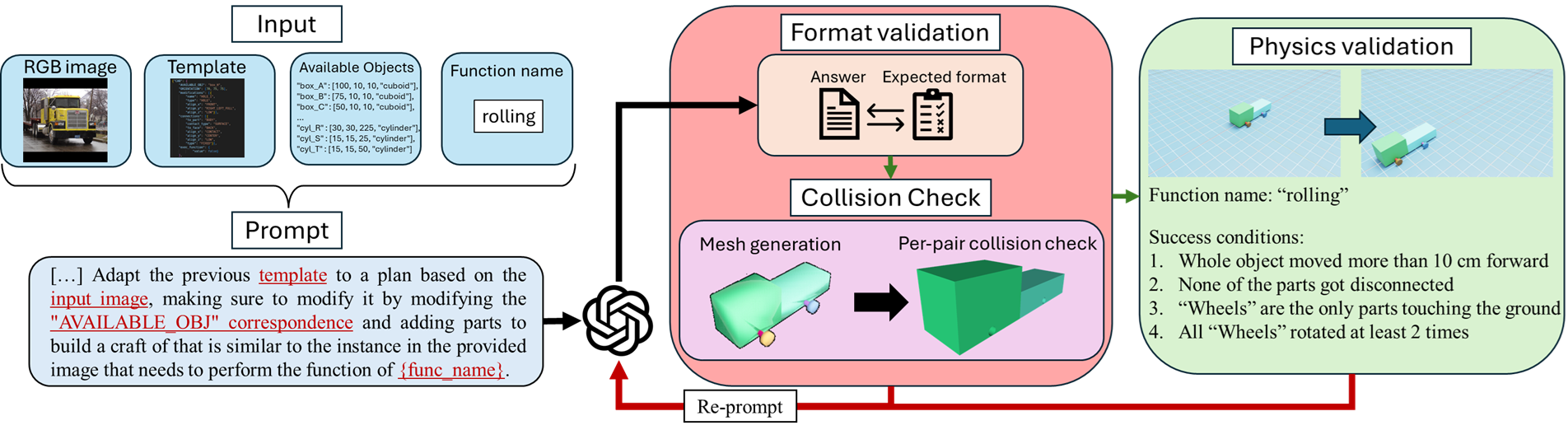}
    \caption{Overview of our approach using LLMs for the Craft Assembly Task. The inputs are used to populate a prompt for the LLM. The response of the LLM goes through multiple checks to generate a successful craft. If the proposal fails at any point, the LLM agent can be re-prompted.}
    \label{fig:overview}
\end{figure*}

\subsection{Problem Setting}
Following the definition of the Craft Assembly Task, we consider the RGB image of a target object in the wild and the textual description of available objects in the scene as the inputs of our proposed system. In this work, we introduce the name of the target function and a structured example file as additional inputs. This file, referred to as a template, provides an example of a successful craft of the target object category using the expected parameter format. The final output of our system is a craft assembly based on the object in the image that is capable of performing the target function. It is defined by a valid structured file that defines the components, their orientations and connectivity relationships.

\subsection{Overview}
Our proposed method is illustrated in Fig.~\ref{fig:overview}. The inputs populate a structured prompt, which is sent to the LLM agent. The response is parsed and verified for format consistency, where validation checks are performed for the parameters names and values. Then, the proposed parts are generated and tested for collision, by positioning and orienting them in 3D space. Here, an additional check is performed to ensure all of them are part of a singular structure. Finally, the craft is generated in a physics simulation, where a function specific test is performed. If a proposed crafts successfully passes through all these steps, it is considered a successful generation. Otherwise, if the proposal failed at any of the validation steps, the LLM agent is re-prompted. 

\subsection{Template format}
To constrain the predictions by the LLM, a template format is designed for the craft assembly, describing each part and its relationships within the craft. The template is a list of parts, with each part entry being composed of:
\begin{enumerate}
    \item Name - A unique name for the part, being composed of the part type name followed by a numeral, to differentiate multiple instances of the same part type; 
    \item Available\_obj - The selected object type from the available set;
    \item Orientation - The alignment of the dimensions along orthogonal axes;
    \item Modifications - Optional boolean operations to create clearance holes / slot, defined by alignment parameters;
    \item Connections - Specifications of inter-part relationships, defined by a contact type (surface contact or insertion), type (fixed or non-fixed) and alignment parameters;
    \item Exec\_function - A binary flag that indicates whether the part needs to perform the designated function or not.
\end{enumerate}

This format enforces a consistent structure for the proposals and bounds the parameter choices, allowing for deterministic parsing into assemblies. The complete description of all fields and allowable values is provided in the Appendix A.

\subsection{Prompt}
The LLM is instructed to adapt the provided template into a feasible assembly plan based on the input image, the available set of objects and the target function. It may modify existing parameters, add new parts, or propose alternative mechanisms for the function implementation as needed. In addition, for each object category, heuristics are included to guide reasoning (e.g., “a hammer must contain one handle and one head”), namely a minimal set of parts and a constraint, though these are not enforced during validation. The full description of the heuristics used are provided in the Appendix B.

\subsection{Available objects in the scene}
The available set of objects in the scene is simulated as a collection of 41 object types, consisting of cuboids and cylinders, with varying dimensions and proportions, ranging from 10 mm to 250 mm along their principal axes. These specifications were chosen arbitrarily to ensure enough diversity of dimensions and proportions is available for a possible solution. The quantity available for each object type is not limited, allowing the agent to choose the same object type for multiple parts.

\subsection{Format Validation}
The response of the LLM is, first, preprocessed to ensure consistency by changing all fields names to uppercase and that hyphens ("-") are replaced with underscores ("\_"). Then, it is parsed to verify if it follows the set guidelines and template format, by checking if all required parameters are present and their values are valid.

\subsection{Collision Validation}
Valid responses are instantiated as 3D assemblies, where each part is generated, oriented and positioned according to the predicted parameters. A collision check is performed to all possible pairs of parts. And, we verify if all parts form a single connected assembly graph, to avoid errors where two or more disconnected sub-assemblies are generated.

\subsection{Physics simulation}
A collision-free proposal does not guarantee that the craft can perform the designated function. Therefore, each proposal is tested in a physics simulator. For stability in the physics simulation, the dimensions of the craft is scaled up by a factor of 10, with all parts being considered as rigid solids with uniform mass of 10 kg and a friction coefficient of 0.5. Parts connected via a fixed connection are attached with fixed joints. Three possible functions are considered: "rolling", "support" and "hit", with different tests and success conditions for each. They do, however, share two failure conditions: if any of the parts separate from the craft during simulation; and if any part that didn't have initial contact with the ground, touches the ground plane during simulation.

For the "rolling" test, a forward force is applied to the part with most connections, typically the main body. Success requires the parts assigned with the flag for function execution to execute at least one full rotation during simulation time, and for the craft to move, at least, one meter forward and not veer to the side.

For the "support" test, a downward force is continuously applied to the surfaces assigned with the function flag. Success requires the parts and the whole craft to remain stable, moving less than 1 cm of distance during the simulation.

For the "hit" test, a peg-and-hole setup if prepared. The peg is held directly above the hole, suspended without the influence of gravity, and the craft is positioned such that the part with the "hit" function is directly above it. The craft is then moved downwards in the Z-axis, and if the "hit" part manages to hit the peg into the hole, it is considered a success, otherwise, if it fails to hit the peg or the peg falls outside of the hole, it is deemed as a failure.

\subsection{Re-prompting}
A typical technique to handle the error due to the stochastic nature of LLMs output is re-prompting. In this work, we evaluate two typical strategies to address it: in the first approach, the system keeps the history of the failed generation and provides structure feedback to the LLM, with the reason of failure and specific information of the validation step where it failed, and asking it to create a new plan, from scratch, while avoiding the errors from the previous plan. In this approach, we consider re-prompting once for plans that failed to reach the physics simulation step, and then once more for plans that failed during the physics simulation, allowing, potentially, two re-prompts. Empirically, in early tests, we did not see significant improvements by allowing more rounds of re-prompting.

In the second approach, the agent is re-invoked with the same initial input and no feedback or information about the previous failure. We also allow for a maximum of two re-prompts, but here we do not restrict it according to when it failed as no feedback is provided.

\section{Evaluation}

\subsection{Implementation details}
For the evaluation of our method, we consider the four object classes in~\cite{isume2024component}- bus, truck, chair and table - and add four new classes: bookshelf, skateboard, scooter and hammer, for more diverse structures. We collect 100 images per object category from ImageNet~\cite{Deng2009} and copyright free sources, for a total of 800 images. We select images where the object is not obstructed, and where a single instance is in focus. One structured template of a valid craft, without a corresponding image, is prepared per object class.

In the input, function labels are assigned per object class, with the "hit" function being assigned to all images of hammer; the "support" function is assigned to all instances of table, chair and bookshelves; and "rolling" is assigned to all images of truck, bus, skateboard and scooter.

The LLM used is OpenAI's o4-mini-2025-04-16, with reasoning effort set to "high", and all other hyperparameters left at their default values. It was selected due to its strong reasoning performance and accepting multi-modal inputs.

The physics-based validation is conducted on NVIDIA's Isaac Sim, with the physics scene configured with a timestep of 500 Hz. All other physics parameters are left to their default values.

\begin{figure*}[t]
    \smallskip
    \centering
    \includegraphics[width=\linewidth]{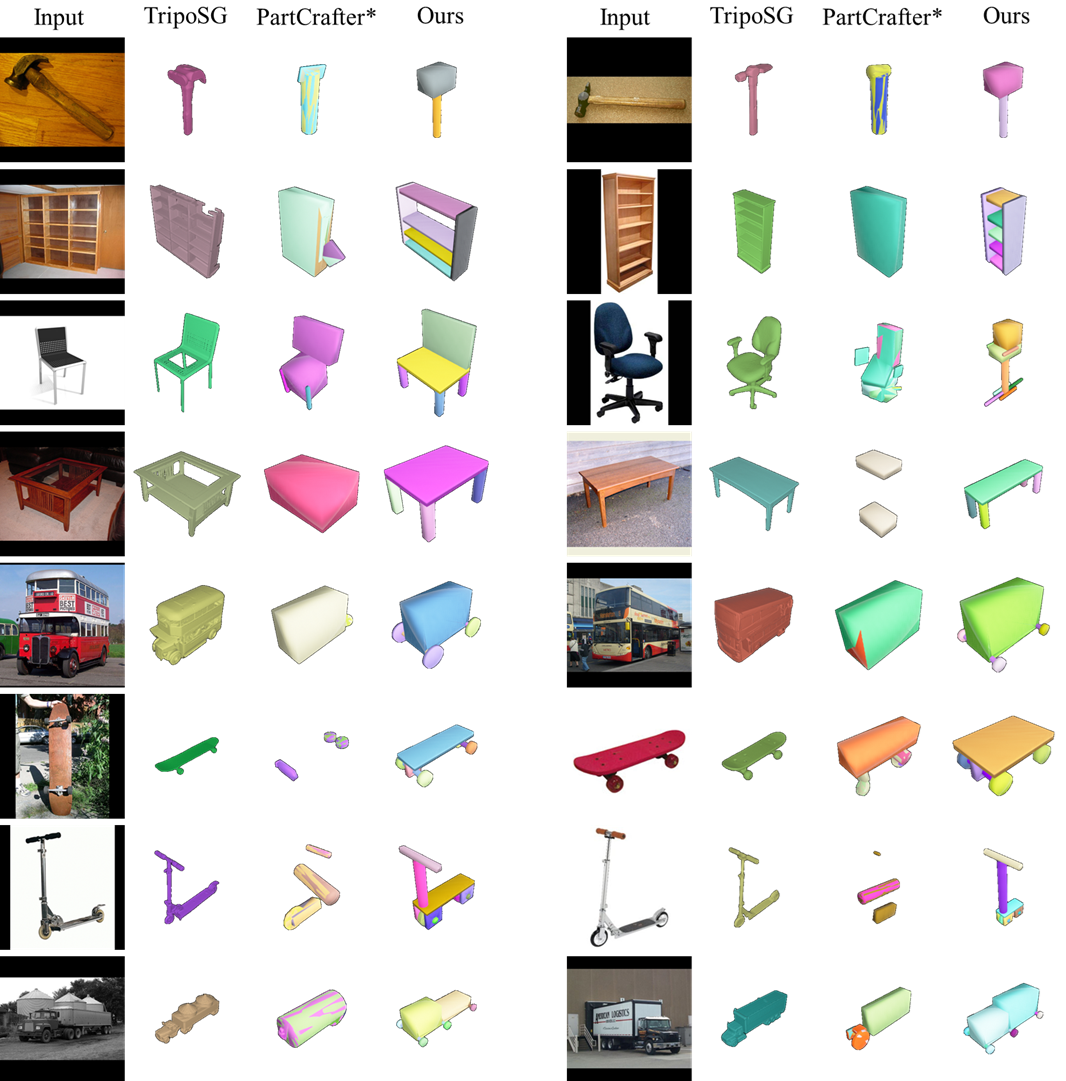}
    \caption{Qualitative results for the generated crafts. TripoSG directly infers the model from the masked image, while the PartCrafter* baseline and Ours (re-prompt w/o feedback) obtain a part separated model and chooses corresponding objects available in the scene. Ours generated more structured results.}
    \label{fig:qualitative}
\end{figure*}

\subsection{Metrics}
In order to evaluate the performance of our method, we split the evaluation into two parts: (1) functional success and (2) visual similarity to the input.

\subsubsection{Functional success}
A proposed craft is deemed a successful generation if it passes all validation steps, including format validation, collision checking and physics simulation. It implies that the proposed structure is valid and capable of executing the target function, measuring the capability of the proposed pipeline to produce a functional craft.

\subsubsection{Visual Similarity}
However, functional validity does not ensure visual similarity to the input image. Since the collected images do not have a ground truth 3D model, we first generate approximate 3D ground truth models for each input image using TripoSG~\cite{li2025triposg}. The resulting meshes are manually refined to remove noise and floating artifacts, and are re-oriented into a canonical pose. These serve as proxy ground truth models for all comparisons.

We also construct a baseline by adapting PartCrafter~\cite{lin2025partcrafter} to our setting. Specifically, we manually segment each image using SAM2~\cite{ravi2024sam2} to mask the object and remove the background, and generate part-separated 3D models using PartCrafter. The number of parts is obtained by only considering the distinct part types in the response of the LLM to the prompt used in our method. The predicted meshes are processed to remove any small volume below an occupancy threshold of $10^{-5}$ of their bounding box volume, and separate non-connected components into independent instances. If this filtering removes the entire mesh, we relax the condition to retain all parts with at least 100 vertices.

The algorithm from \cite{isume2024component} is applied to simplify all generated parts to primitive shapes and select the subset of available objects that best match the primitive shapes. It is important to note that this baseline prioritizes visual resemblance and does not enforce functional correctness or collision constraints.

For numerical evaluation, we uniformly sample 100,000 points on the exterior surface of the each mesh, and compute the chamfer distance, Hausdorff distance and F-score at 0.1 distance threshold to evaluate the shape similarity of our method and the baseline compared to the generated models from TripoSG. Since they do not contain internal volumes, all interior vertices and faces in the meshes generated by our pipeline and the baseline are ignored during sampling.

\subsection{Craft generation results}
We report the successful generations for four variants of our approach, analyzing the effect of re-prompting and feedback strategies in Table~\ref{tab:ablation}. The initial LLM response without any re-prompting or feedback, achieved an overall 63.4\% success rate (w/o Re-prompt). Re-prompting failed cases up to two times without feedback, improved performance significantly to 85.0\% (Re-prompt). Re-prompting one time for failed cases before the simulation and providing explicit feedback for collision and format errors resulted in a 75.7\% success rate (+Collision). Re-prompting once more for failed cases in the physics simulation, with explicit feedback further increases the success rate to 77.1\% (+Collison+Simulation). 

Interestingly, re-prompting from scratch yielded significantly better results than explicitly guiding the model to fix specific errors. One reason for this difference is that most errors stems from the collision validation step, and for a failed generation, the re-prompt without feedback approach could retry two times, while in the feedback approach, if it failed to generate a collision-free proposal after the first re-prompt, it would not be re-prompted in the second one. We also hypothesize that once an LLM has proposed an invalid design, subsequent prompts often remain constrained by the initial reasoning, leading to minor or ineffective modifications, even though in the re-prompt, we ask it to generate a new plan and to only be mindful of the previous errors. This behavior may also have been caused by the limited control over the model parameters as stochastic sampling in the LLM can yield remarkably different results from the same prompt, making it more effective to simply sample multiple answers.

In regards to performance per object class, we observed a noticeable gap between bookshelves, chair and scooter compared to the other object classes. For bookshelves, the LLM may position multiple shelves at the same location or fails to consider the collision of possible vertical dividers with the horizontal shelves. The chair category contains a wide variety of shapes and structures, with designs ranging from a simple four-legged chair with a backrest to a office chair with a five-star base, as shown in Fig.~\ref{fig:qualitative}. Although the LLM demonstrated adaptability in transforming the template to such cases, it also often failed to account for collisions and placement of the newly proposed parts. Similarly, for the scooter, which require tighter tolerances for wheel clearance, most errors stems from collisions between parts.

\begin{table}[t]
    \smallskip
    \smallskip
    \centering
    \caption{Ablation study for different re-prompt techniques. Relative increments compared to initial results shown in parenthesis.}
    \resizebox{\linewidth}{!}{%
    \begin{tabular}{l| c l | l  l}
    \multicolumn{1}{c|}{} & \multicolumn{2}{c|}{w/o feedback}  & \multicolumn{2}{c}{w/ feedback} \\
    Class & w/o Re-prompt & Re-prompt & + Collision &  + Collision + Simulation.   \\
    \hline
    Hammer      & 81\% & \hspace{0.5em}95\% (+14) & \hspace{1em}84\% (+3)  & \hspace{3.5em}86\% (+5)\\
    Shelves     & 53\% & \hspace{0.5em}80\% (+27) & \hspace{1em}73\% (+20) & \hspace{3.5em}73\% (+20)\\
    Chair       & 49\% & \hspace{0.5em}71\% (+22) & \hspace{1em}64\% (+15) & \hspace{3.5em}66\% (+17)\\
    Table       & 76\% & \hspace{0.5em}83\% (+7)  & \hspace{1em}81\% (+5)  & \hspace{3.5em}81\% (+5)\\
    Bus         & 93\% & \hspace{0.1em}100\% (+7) & \hspace{1em}95\% (+2)  & \hspace{3.5em}96\% (+2)\\
    Scooter     & 35\% & \hspace{0.5em}63\% (+28) & \hspace{1em}50\% (+15) & \hspace{3.5em}53\% (+18)\\
    Skateboard  & 61\% & \hspace{0.5em}94\% (+33) & \hspace{1em}84\% (+23) & \hspace{3.5em}85\% (+24) \\
    Truck       & 59\% & \hspace{0.5em}95\% (+36) & \hspace{1em}75\% (+16) & \hspace{3.5em}77\% (+18) \\
    \hline
    \textbf{Overall} & 63.375\% & \hspace{1.5em}85.0\% & \multicolumn{1}{c}{75.75\%} & \multicolumn{1}{c}{77.125\%}
    \end{tabular}
    }
    \label{tab:ablation}
\end{table}

\subsection{Visual similarity}
Quantitative results comparing functionally successful cases of the best performing variant of our method and the adapted PartCrafter baseline, denoted with an asterisk, are shown in Table~\ref{tab:success}. PartCrafter, which is built upon TripoSG is expected to produce meshes with higher similarity to the TripoSG-generated references. However, we noticed that the generated parts often contained residual noise or surface deformations, which lead to overlap or inaccurate measurements when simplifying it to primitive fitting. To mitigate this, we applied the same filtering criteria we used for TripoSG. The generation of parts by PartCrafter also failed sometimes, hallucinating parts, possibly due to the low quality images compared to the training data used in their model. These issues can be observed in Fig.~\ref{fig:qualitative} for the skateboard, scooter and table. Nonetheless, the baseline tends to overfit to the generated mesh since collision and functionality are ignored, and they are manually re-oriented to match the canonical orientation of the objects.

While the baseline achieved favorable performance for large, box-shaped objects (e.g., bookshelves and bus), our method outperformed it in objects with complex geometry or thin profiles such as the chair, scooter, skateboard and truck, reflecting its ability to maintain geometric coherence under functional constraints.

Notable deviations are observed in the hammer and table categories. In several hammer instances, our method positioned the head in the midpoint of the handle rather than at the end, which significantly reduced the visual similarity scores despite being functionally valid. For the table, the biggest impact comes from the output of our method often aligning the largest edge of the table along the X-axis, as compared to the reference, which is placed along the Y-axis. Across all objects, the performance in all three metrics was mostly consistent, and qualitative results indicate that the crafts generated by our method are visually coherent and structurally plausible (Fig.~\ref{fig:qualitative}).

\begin{table}[t]
    \smallskip
    \smallskip
    \centering
    \caption{Visual similarity metrics compared to meshes generated by TripoSG.}
    \resizebox{\linewidth}{!}{%
    \begin{tabular}{l| c c | c c | c c}
     & \multicolumn{2}{c|}{Chamfer Distance $\downarrow$} & \multicolumn{2}{c|}{Hausdorff Distance $\downarrow$}
    & \multicolumn{2}{c}{F-Score@0.1 $\uparrow$}\\
    \hline
    Class & PartCrafter* & Ours & PartCrafter* & Ours & PartCrafter* & Ours  \\
    \hline
    Hammer      & \textbf{0.444} & 0.536 & \textbf{0.669} & 0.764 & 0.291 & \textbf{0.325} \\
    Shelves   & \textbf{0.301} & 0.344 & \textbf{0.519} & 0.602 & \textbf{0.408} & 0.395\\
    Chair       & 0.408 & \textbf{0.389} & 0.777 & \textbf{0.689} & \textbf{0.317} & 0.312 \\
    Table       & \textbf{0.502} & 0.633 & \textbf{0.772} & 0.775 & \textbf{0.306} & 0.162\\
    Bus         & \textbf{0.251} & 0.403 & \textbf{0.414} & 0.623 & \textbf{0.481} & 0.217\\
    Scooter     & 0.536 & \textbf{0.306} & 0.772 & \textbf{0.580} & 0.251 & \textbf{0.414}\\
    Skateboard  & 0.523 & \textbf{0.369} & 0.757 & \textbf{0.630} & \textbf{0.269} & 0.257\\
    Truck       & 0.252 & \textbf{0.211} & 0.457 & \textbf{0.392} & 0.515 & \textbf{0.537}\\
    \hline
    \textbf{Average} & \textbf{0.395} & 0.401 & \textbf{0.630} & 0.630 & \textbf{0.360} & 0.325 
    \end{tabular}
    }
    \label{tab:success}
\end{table}

\begin{table}[tb]
    \centering
    \caption{Failures categorized by validation step.}
    \resizebox{\linewidth}{!}{%
    \begin{tabular}{l| c c c | c c c}
    \multicolumn{1}{c|}{} & \multicolumn{3}{c|}{Initial generation
    }  & \multicolumn{3}{c}{After 2nd re-prompt w/o feedback} \\
    Class & Format Val. & Position Val. & Physics Val. & Format Val. & Position Val. & Physics Val.   \\
    \hline
    Hammer      & 6 & 9 & 4 & 4 & 1 & 0 \\
    Shelves     & 17 & 30 & 0 & 8 & 12 & 0 \\
    Chair       & 18 & 28 & 5 & 9 & 20 & 0 \\
    Table       & 3 & 21 & 0 & 3 & 15 & 0 \\
    Bus         & 0 & 3 & 4 & 0 & 0 & 0\\
    Scooter     & 8 & 45 & 12 & 7 & 26 & 4 \\
    Skateboard  & 4 & 34 & 1 & 0 & 5 & 1 \\
    Truck       & 2 & 30 & 9 & 1 & 4 & 0 \\
    \hline
    \textbf{Total} & 58 & 200 & 35 & 32 & 83 & 5
    \end{tabular}
    }
    \label{tab:failure}
\end{table}

\subsection{Failure cases}
\begin{figure}
    \smallskip
    \centering
    \includegraphics[width=0.9\columnwidth]{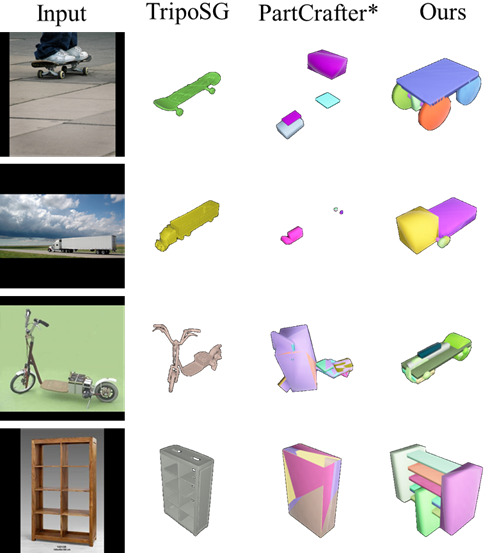}
    \caption{Failure cases of our method. Most of the failures are caused by collision between parts.}
    \label{fig:failure}
\end{figure}

We report the causes of failure in the initial generation, and after the two re-prompt steps in Table~\ref{tab:failure}. The main cause of failure is in the collision validation step, accounting for 69.1\% of the errors. This result aligns with the prior reported limitations of LLMs in 3D spatial reasoning. In particular, the model struggles to account for possible obstructions between parts that are not directly connected. Other common errors include misplacement and incorrect orientation of parts. These errors can be visually observed in Fig.~\ref{fig:failure}.

Errors during the format validation step are mostly caused by invalid or incomplete parameters, such as hallucinated dimensions for cuboids or missing alignment fields. 

Finally, in the case of physics simulation, failures typically occur due to misplacement of parts, leading to a craft that is functionally incorrect. For example a scooter's wheels not touching the ground, or wheels not being able to rotate properly due to being misaligned with the center of the axles.

To mitigate such issues, fine-tuning LLMs with explicit 3D spatial reasoning or incorporating physics-aware foundation models~\cite{agarwal2025cosmos} could improve performance. In the case of the scooter or skateboard, introducing a validation module for clearance between non-connected parts or autonomously refining the constraints in the prompt through another LLM agent could improve the results while keeping robustness. We leave exploration of these approaches to future work.

\section{Conclusions}
In this work, we introduced a novel framework that leverage LLMs as decision making agents for the Craft Assembly Task. This approach addresses limitations in scalability and robustness from previous work. Additionally, we expand the craft assemblies to include functionality implementation and validate them through physics simulation tests. The proposed method achieved a high success rate in generating valid and functional crafts, and in visual similarity, it performed comparably or better than a baseline using image-to-3D parts model. However, this could be further improved in future work by using a separate critic agent to iterate through different designs to ensure better visual similarity, or adding more modifications to the selected objects to further customize it.

We also recognize that although our method further decreases the need of expert knowledge or prior data, it still relies on one handcrafted template for the object category. The physics simulation test are also designed per function, and although they are  designed while prioritizing simplicity, a possible future direction is further modularizing them, so they can be generated through additional LLM agents or through foundation models with physics understanding.

\section*{Appendix}

\subsection{Template Format}
The customized JSON file format is represented in Fig.~\ref{fig:template}. In the prompt, an example of a craft assembly of the target object category is provided using this format, and in the output configuration, the LLM is instructed to output a list of parts with this format to describe the craft assembly. The following coordinate system is assumed for the environment:

\begin{itemize}
    \item X-axis corresponds to the "FRONT-BACK" orientation;
    \item Y-axis corresponds to "RIGHT-LEFT" orientation;
    \item Z-axis corresponds to "TOP-BOTTOM" orientation;
\end{itemize}

\subsection{Object-specific heuristics}
Alongside each template, an additional set of heuristics is provided for each object category. Namely, a list of minimal parts and a short description of a constraint, as shown in Fig.~\ref{fig:minimal} and Fig.~\ref{fig:constraint}, respectively.

The list of minimal parts is the list of the parts from each template, while the constraints were generated from the most common error observed in early empirical testing. These conditions, however, are not enforced during validation.

\begin{figure}[t]
    \smallskip
    \centering
    \begin{lstlisting}[language=json]
{
"Name": Part type name + _ + numeral,
"Available_obj": Chosen object type from provided list
"Orientation": { 
    If "Available_obj" is a cuboid: [dim_x, dim_y, dim_z]
    If "Available_obj" is a cylinder: one of ["FRONT_BACK", "LEFT_RIGHT", "TOP_BOTTOM"]. Indicates the alignment of the principal axis},
"Modifications": [{
    "Name": Modification type + _ + numeral
    "Type": One of ["HOLE"]. Currently only "HOLE" is considered.
    "Align_x": If the modification is not along x-axis, one of ["FRONT", "CENTER", "BACK"]
    If the modification is along x-axis, one of ["FRONT_BACK_FULL", "FRONT_BACK_HALF", "BACK_FRONT_HALF"]. Indicating the direction and legth of the modification.
    "Align_y": If the modification is not along y-axis, one of ["RIGHT", "CENTER", "LEFT"]
    If the modification is along y-axis, one of ["RIGHT_LEFT_FULL", "RIGHT_LEFT_HALF", "LEFT_RIGHT_HALF"]. Indicating the direction and legth of the modification.
    "Align_z": If the modification is not along z-axis, one of ["HIGH", "CENTER", "LOW"]
    If the modification is along z-axis, one of ["HIGH_LOW_FULL", "HIGH_LOW_HALF", "LOW_HIGH_HALF"]. Indicating the direction and legth of the modification.}],
"Connections": [{
    "to_part": Name of the connected part,
    "contact_type": One of ["Surface", "Inserted"],
    "to_face": If "contact_type" is "Surface", one of ["TOP", "BOTTOM", "RIGHT", "LEFT", "FRONT", "BACK"]. Defines where the connection occurs in the the current part.,
    "align_x": If "contact_type" is "Surface", one of ["FRONT", "CENTER", "BACK"],
    "align_y": If "contact_type" is "Surface", one of ["RIGHT", "CENTER", "LEFT"],
    "align_z": If "contact_type" is "Surface", one of ["TOP", "CENTER", "BOTTOM"],
    "to_modification": If "contact_type" is "Inserted", the name of the modification where the part needs to be inserted.}],
"exec_function": One of [true, false]. Indicates whether the part performs the given function or not}
    \end{lstlisting}
    \caption{JSON template format for one part of the assembly.}
    \label{fig:template}
\end{figure}

\begin{figure}[!h]
    \smallskip
    \centering
    \begin{lstlisting}[language=json]
    "Hammer": {
        "Head": 1,  "Handle": 1},
    "Bookshelf": {
        "Side_Panel": 2,  "Shelf": 3},
    "Chair": {
        "Seat": 1,  "Leg": 4},
    "Table": {
        "Tabletop": 1,  "Leg": 4},
    "Bus": {
        "Body": 1,  "Wheel": 4,  "Axle": 2},
    "Scooter": {
        "Deck": 1,  "Stem": 1,  "Handlebar": 1,
        "Support": 4,  "Wheel": 2,  "Axle": 2},
    "Skateboard": {
        "Deck": 1,  "Support": 2,  "Wheel": 4,
        "Axle": 2}
    \end{lstlisting}
    \caption{List of minimal set of parts}
    \label{fig:minimal}
\end{figure}

\begin{figure}[!h]
    \centering
    \begin{lstlisting}[language=json]
    "Hammer": "The HEAD is typically placed towards the top of the object.",
    "Bookshelf": "Each SHELF must be connected to two side panels."
    "Chair": "Ensure the position and orientation of the parts is correct.",
    "Table": "Ensure the position and orientation of the parts is correct.",
    "Bus": "For the AXLES and WHEELS, the flat surfaces must be in contact to be a valid SURFACE connection. To achieve this, they must follow the same ORIENTATION and they must be centered.",
    "Scooter": "For the AXLES and WHEELS, they must follow the same ORIENTATION and they must be centered. The WHEELs width must be within the space between the supports for the axles, and must not collide with the DECK.",
    "Skateboard": "For the AXLES and WHEELS, they must follow the same ORIENTATION and they must be centered. The WHEELs must not collide with the DECK.",
    "Truck": "For the AXLES and WHEELS, the flat surfaces must be in contact to be a valid SURFACE connection. To achieve this, they must follow the same ORIENTATION and they must be centered."
    \end{lstlisting}
    \caption{List of constraints per object category}
    \label{fig:constraint}
\end{figure}






\bibliographystyle{IEEEtran}
\bibliography{IEEEabrv, all_ref.bib}

\end{document}